\documentclass{article}
\usepackage{log_2024}	

\usepackage[utf8]{inputenc} 
\usepackage[T1]{fontenc}    
\usepackage{url}            

\usepackage{booktabs}						
\usepackage{multirow}						
\usepackage{amsfonts}						
\usepackage{graphicx}						
\usepackage{duckuments}						

\usepackage{amsfonts}       
\usepackage{nicefrac}       
\usepackage{microtype}      
\usepackage{xcolor}         

\usepackage{wrapfig}

\usepackage{amsmath,amsfonts,bm}

\def\eqref#1{equation~\ref{#1}}

\def\1{\bm{1}}

\DeclareMathAlphabet{\mathsfit}{\encodingdefault}{\sfdefault}{m}{sl}
\SetMathAlphabet{\mathsfit}{bold}{\encodingdefault}{\sfdefault}{bx}{n}

\DeclareMathOperator*{\argmin}{arg\,min}

\usepackage{arash_macros2}

\usepackage{subcaption,booktabs}
\usepackage{siunitx}

\usepackage{enumitem}

\renewcommand{\eqref}[1]{(\ref{#1})}

\newcommand{\numwithuncertainty}[2]{#1{\scriptsize\,$\pm$\,#2}}

\usepackage[numbers,compress,sort]{natbib}

\title[Simple GNNs with Low Rank Non-parametric Aggregators]{Simple GNNs with Low Rank Non-parametric Aggregators}

\author[L. Vinas et al.]{%
Luciano Vinas \and Arash A. Amini\\
University of California, Los Angeles\\
\email{lucianovinas@g.ucla.edu}}

\begin{document}

\maketitle

\begin{abstract}
We revisit recent spectral GNN approaches to semi-supervised node classification (SSNC). We posit that state-of-the-art (SOTA) GNN architectures may be over-engineered for common SSNC benchmark datasets (citation networks, page-page networks, etc.). By replacing feature aggregation with a non-parametric learner we are able to streamline the GNN design process and avoid many of the engineering complexities associated with SOTA hyperparameter selection (GNN depth, non-linearity choice, feature dropout probability, etc.). Our empirical experiments suggest conventional methods such as non-parametric regression are well suited for semi-supervised learning on sparse, directed networks and a variety of other graph types commonly found in SSNC benchmarks. Additionally, we bring attention to recent changes in evaluation conventions for SSNC benchmarking and how this may have partially contributed to rising performances over time.
\end{abstract}

\section{Introduction}

The problem of semi-supervised node classification (SSNC)~\citep{Seeger02,Belkin06} has been a focal point in graph-based classification for roughly 20 years. At the task's inception, classical methods such as label propagation~\cite{Zhu03} and kernel learning~\cite{Belkin06} had seen moderate success in predicting unobseved node labels. Now, in an era where computation is more plentiful, modern approaches to the classification problem on graphs make use of the multilayer Graph Neural Network (GNNs)~\citep{Scarselli09}.

These networks, trained to predict node labels in SSNC, draw on both the individual node features ($\bm X$) and the broader network structure ($\bm A$) to inform their prediction.

The fundamental premise of SSNC is that the network structure allows us to borrow information from neighboring nodes for which we lack a response. This borrowing can enhance the prediction of the unobserved responses ($\bm y$) beyond what could be achieved with a traditional regression solely on node features. Recently, there has been a wide breadth of literature~\citep{Velickovic18,Chien21,Luan22} which attempts to better leverage the network structure of the graph using GNNs. This recent flurry of activity has led to the proposal of many competing, and often intricate, architectures to solve the SSNC problem.

Our study of the leading GNN architectures and the benchmarks used to prove their algorithmic effectiveness, has led us to believe that many of the design choices found in modern GNNs may be drastically simplified, or even removed completely, at little-to-no cost to predictive performance. In our efforts to validate model performances, we revisit traditional estimation techniques like non-parametric regression. These techniques happen to be very effective for SSNC and highlight the importance of learnable feature aggregation in SSNC problems. 

To this end, we devise a flexible non-parametric learner for feature aggregation. This learner generalizes the specific polynomial form used in spectral GNNs~\cite{Defferrard16,Wang22}. That is, given a singular value decomposition of the network graph $\bm{A} = \bm{U} \bm{\Sigma} \bm{V}^T$, the non-parametric learner $f:\reals\to\reals$ transforms the spectrum of $\bm{A}$ to produce a new aggregation matrix 
\[
\bm{P}_{f} = \bm{U} f(\bm{\Sigma}) \bm{V}^T
\]
where $f$ is applied entry-wise across the diagonal of $\bm{\Sigma}$. This singular value extension to the previous symmetric spectral approach of~\cite{Wang22} helps clear a directed graph hurdle faced by previous spectral GNN techniques.

Our contributions are as follows: 
\begin{enumerate}
[wide, labelwidth=!, labelindent=0pt, left=0.5cm]
    \item Propose a nonparametric approach to learn $f$, hence a GNN aggregation operator, by borrowing ideas from the theory of reproducing kernel Hilbert spaces (RKHS), thus generalizing polynomial aggregation to a much broader class of spectral functions. By controlling the underlying kernel,  one can impose different regularity constraints on the spectral filters. 
   
    \item Highlight the importance and sensitivity of nonparametric spectral reshaping and show how it can be used to simpify model hyperparameters (e.g. dropout probabilities, model depth, parameter-specific optimizers) at near-no-cost to SOTA performance.
    \item Classification improvements of +5\% and +20\% compared to competing spectral methods and other non-linear GNN baselines for the challenging benchmark datasets Chameleon and Squirrel~\citep{Rozemberczki21}.

     \item Outline common evaluation practices which have an outsized effect on model performance.
\end{enumerate}
By standardizing evaluation practices and simplifying modeling considerations, we aim to disambiguate performance in the GNN model-space and hope to encourage more interpretable models and heuristics for future SSNC problems.

\section{GNN and SSNC Formalism}

In our observation framework, we consider observing a, potentially noisy realization, of the network with adjacency matrix $\bm A\in\mathbb{R}^{n\times n}$ and node feature matrix $\bm{X} \in \mathbb{R}^{n\times d}$. Specifically, each node in the network $i\in[n]$ is associated with a feature vector $\bm{x}_i$ and a label $y_i \in [C] := \{1,\dots,C\}$.

In SSNC, it is assumed that for a subset of nodes $\mathcal{O}\subset[n]$ the labels $(y_i)_{i\in\mathcal{O}}$ are observed. In this setting, both the adjacency matrix $\bm{A}$ and the feature matrix $\bm{X}$ are assumed to be fully observed. The goal then is to correctly predict unobserved labels $(y_i)_{i\in\mathcal{O}^c}$ from the previously stated knowns.

GNNs are designed layerwise, with non-linearity $\phi^\ell:\reals\to\reals$, weight matrix $\bm{W}^\ell\in\reals^{d_{\ell}\times d_{\ell-1}}$ and aggregation matrix $\bm{P}^{\ell}\in\mathbb{R}^{n\times n}$ all depending on layer $\ell \in [L]$. Placed altogether, the intermediate features of the GNN can be expressed as
\begin{equation}\label{eq:gnn:blueprint}
    \bm{Z}^{\ell+1} = \phi^\ell(\bm{P}^\ell \bm{Z}^\ell \bm{W}^\ell)
\end{equation}
with $\phi^\ell$ applied element-wise, $d_0 = d$ and $\bm{Z}^1 = \bm{X}$. In the case of a $C$-class classification problem, it is common to extract row-wise ``argmax''s 
of the final features $\bm{Z}^L\in\mathbb{R}^{n\times C}$ using differentiable argmax surrogates
such as ${\rm softmax}$. Choice of the aggregation matrix $\bm{P}^\ell$ may vary dramatically depending on architecture, but common choices include the adjacency matrix $\bm A$, its transformed variants (e.g. normalized Laplacian), and other, learnable, attention-based mechanisms~\citep{Velickovic18}.

\subsection{Nonparametric Spectral Reshaping}

In our proposed model, we consider the simplest variant of GNN: a one layer ($L=1$), linear GNN, that is $\phi = {\rm id}$, where special attention is paid to the propagation structure $\bm P$.
For ease of exposition, we first consider the undirected case where the adjacency matrix $\bm A$ is symmetric. Let $\bm M$ be a (symmetric) \emph{network matrix} derived from $\bm A$. Examples include $\bm M \in \{\bm A, \bm D - \bm A,  \hat{\bm A}, \bm I - \hat{\bm A}\}$ where $\hat{\bm A} = \bm D^{-1/2} \bm A \bm D^{-1/2}$. Our approach is to consider a general nonlinear deformation of $\bm M$, namely, $f(\bm M)$ where $f: \reals \to \reals$ is a univariate function extended to the space of symmetric matrices by the so-called \emph{functional calculus}. More precisely, given the eigendecomposition $\bm M = \bm U \bm \Lambda \bm U^T$ of the $\bm M$ matrix, where $\bm \Lambda = \diag(\lambda_i, i \in [n])$, one has 
\[
f(\bm M) = \bm U f(\bm \Lambda) \bm U^T
\]
where $f(\bm \Lambda) = \diag(f(\lambda_i), i \in [n])$ is the natural extension of $f$ to diagonal matrices. This way of extending univariate functions to self-adjoint operators has a long history in operator theory. Thus, our propagation operator is $\bm P_f = f(\bm M)$ and we propose to optimize a loss over a general class of functions $\mathcal F$:
\begin{align}\label{eq:f:optim:1}
    \hat f = \argmin_{f \in \mathcal F,\; \bm W \in \reals^{d \times C}} \sum_{i \in \mathcal O} \ell\big(y_i, (f(\bm M) \bm X \bm W)_i\big) + \text{pen}(f)
\end{align}
where $\text{pen}(f)$ is some regularization penalty on $f$. Our main claim is that rather than assuming a specific parametric form for $f$, one can allow $f$ to range in a potentially infinite-dimensional function space $\mathcal F$.

Of particular interest to us is when $\mathcal F = \mathbb H$, a reproducing kernel Hilbert space (RKHS) of functions, characterized by a  kernel function $\mathcal{K}:\reals\times\reals\to\reals$. In such a space, the Hilbert norm $\|f\|_{\mathbb H}$ measures irregularity of $f$. Then, as long as $\text{pen}(f)$ is a monotonic function of the Hilbert norm $\|f\|_{\mathbb H}$, by the so-called represented theorem~\citep{Scholkopf01},  problem~\eqref{eq:f:optim:1} reduces to
\begin{align}\label{eq:alpha:optim:1}
    \hat{\bm \alpha} &=\argmin_{\bm \alpha \in \reals^n,\, \bm W} 
    \sum_{i \in \mathcal O} \ell\big(y_i, (\bm P_{\mathcal K}(\bm \alpha) \bm X \bm W)_i\big) +  \widetilde{\text{pen}}(\bm \alpha), \\
    &\text{where}\quad \bm{P}_{\mathcal{K}}(\bm \alpha) := \bm{U}(\diag(\bm{K}\bm{\alpha})) \bm{U}^T,
    \label{eq:prop:K:alpha}
\end{align}
and $\bm K \in \reals^{n \times n}$ is the kernel matrix with entries $K_{ij} = \mathcal K(\lambda_i, \lambda_j)$. If $\text{pen}(f) = \omega(\|f\|_{\mathbb H})$ for monotonic function $\omega: \reals_+ \to \reals$, then $\widetilde{\text{pen}}(\bm \alpha) = \omega(\bm \alpha^T \bm K \bm \alpha)$. Given $\hat{\bm \alpha}$ one can explicitly write down the solution $\hat f$ of the functional problem~\eqref{eq:f:optim:1} as
\[
\hat f(\lambda) := \sum_{j} \hat{\bm\alpha}_j \mathcal K(\lambda, \lambda_j)
\]
which is the learned spectral filter.

\paragraph{Practical considerations.}

We found slight improvements in performance when 
regularizing with $\bm \alpha^T \bm \alpha$ rather than the Hilbert norm surrogate ($\bm \alpha^T \bm K \bm \alpha$). This amounts to using $\widetilde{\text{pen}}(\bm \alpha) = \rho\,  \bm \alpha^T \bm \alpha$ for some $\rho > 0$. When minimizing with GD type methods, this is equivalent to introducing weight decay $\rho$, and is already built into SOTA solvers.

Additionally, we consider the possibility that 
edges in the network themselves have a component of randomness associated with them (Section~\ref{sec:motivate}).

This means that our initial spectral inputs $\lambda_1,\lambda_2,\ldots,\lambda_n$ are themselves noisy. It is then natural to truncate the spectral decomposition of $\bm M$ to the top $r$ eigenvalues (in absolute values). Thus, if the eigenvalues are ordered as $|\lambda_1| \ge |\lambda_2| \ge \cdots \ge |\lambda_n|$, we consider $\bm M^{(r)} = \bm U \bm \Lambda^{(r)} \bm U^T$ where $\bm \Lambda^{(r)} = (\lambda_i, i \in [r])$ and let the aggregation matrix be $f(\bm M^{(r)}) = \bm U f( \bm \Lambda^{(r)} ) \bm U^T$. Following through as before, the only changes to the algorithm is to replace $\bm K$ in~\eqref{eq:prop:K:alpha} with $\bm K^{(r)} = (\mathcal K(\lambda_i, \lambda_j))_{i,j=1}^r$. We also note that $\bm \alpha$, the learnable spectral parameter, will be $r$-dimensional in this case. We treat the $r \in [n]$ as a hyperparameter and study its effect in simulations. We refer to the case $r < n$ as low-rank (LR) kernel model. 

\paragraph{Directed/asymmetric case.} All the above naturally extends to directed networks, where $\bm M$ is not necessarily symmetric, by replacing the eigenvalue decomposition with the SVD: $\bm M = \bm U \bm \Sigma \bm V^T$ where $\bm \Sigma = \diag(\sigma_i, i \in [n])$ collects the singular values of $\bm M$. The aggregation matrix in this case is $
\bm{P}_{f} = \bm{U} f(\bm{\Sigma}) \bm{V}^T
$ and its finite-dimensional version is $\bm{P}_{\mathcal{K}}(\bm \alpha) = \bm{U}(\diag(\bm{K}\bm{\alpha})) \bm{V}^T$ with the kernel matrix $\bm K = (\mathcal K(\sigma_i,\sigma_j))_{i,j=1}^n$ now based on singular values. Everything else follows similarly, including rank truncation, where we use ordered singular values instead. 

\paragraph{Multiple layers.} We mainly focus on a single-layer model (with identity activation) and empirically show that a single layer of this model is enough to achieve near SOTA performance. However, it is straightforward to extend the model to multiple layers via the general blueprint~\eqref{eq:gnn:blueprint} where each layer will have aggregation operator $\bm P^{\ell} = f_\ell(\bm M)$ with $f_\ell$ belonging to $\mathbb H$.

\subsection{Motivating General Spectral Learners}\label{sec:motivate}
Implicit in all graph learning problems is the assumption that node features $\bm X$ are only partially informative towards 
predicting
$\bm y$. 
To motivate why a spectral GNN of the form~\eqref{eq:f:optim:1}, with a general  reshaping function $f$ can improve prediction, let us consider perhaps the simplest theoretical model of SSNC, the so-called Contextual Stochastic Block Model (CSBM)~\citep{Deshpande18}. The idea is that the labels $\bm y$ are latent variables generating both $\bm A$, via a $C$-class SBM: $\pr(A_{ij} = 1 \,|\, \bm y) = B_{y_i, y_j}$, and the node features via a mixture model: $\bm x_i \, |\, y_i \sim N(\bm \mu_{y_i}, \sigma^2 I)$. 

\newpage

\begin{wrapfigure}{r}{0.4\textwidth}
  \centering
  \includegraphics[width=0.4\textwidth]{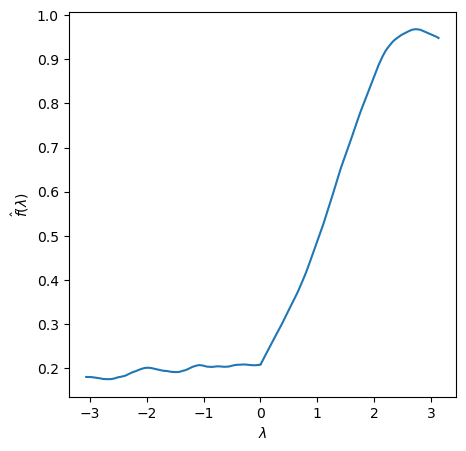}
  \caption{\small Example of learned spectral filter $\hat f(\lambda)$ for the CSBM experiment with an unbounded Sobolev kernel ($\gamma = 0.1$).}
\end{wrapfigure}

In this case, the idealized version of $\bm A$ is $\ex[\bm A]$ which is a rank $C$ matrix with $C$ eigenvectors that are \emph{indicator vectors} of each of the $C$ classes $\Gamma_1,\dots,\Gamma_C \in\{0,1\}^n$. Consequently, if $\bm A = \bm U \bm \Lambda \bm U^T$ is the EVD of $\bm A$, one expects $\bm U_{:j} \approx \bm 1_{\Gamma_j}$ for $j \in [C]$, while $\bm U_{:j}$ for $j > C$ are expected to be mostly noise. So an ideal aggregation operator is close to $f(\bm M) = \bm U f(\bm \Lambda) \bm U^T$ where $f$ is a step function that passes the $\lambda_j, j \in [C]$ through and zero out the rest. As the experiment in Section~\ref{sec:csbm} show, this is mostly what happens when we train~\eqref{eq:f:optim:1} on CSBM, albeit with more nuance. In finite samples, $\bm A$ is not exactly low-rank and lower eigenvectors might still have information about  $\bm y$. This is what we observe in practice where the learned $f$ is a \emph{tapered} thresholding operator, that gradually downweights lower frequencies.

The general low-rank behavior of the $\ex[\bm A]$ is not limited to SBMs and holds for more realistic network models such as random dot product graph (RDPG)~\citep{Young07} where depending on the distribution of latent positions, more complex tapering might be optimal. 

\subsection{Complexity of Low Rank Spectral Learners}

Scalability remains an issue for dense spectral methods. However in the case of low-rank non-parametric aggregators, this issue can be addressed through the use of a low-rank spectral approximation. By first selecting a rank parameter $r$ for the non-parametric aggregator, computation can be better budgeted ahead of time for a graph $G = (V,E)$ through the use of a low-rank SVD approximations. Specifically for PyTorch, a low-rank, random SVD routine based on~\cite{Halko11} is implemented in the function \texttt{torch.svd\_lowrank}. 

Computation and error complexity for this routine can be found in section 6.2 of~\cite{Halko11}. For a sparse adjacency matrix given by $G$ and a number of total iterations $q$, this routine has a time complexity of $\mathcal{O}(qr|E|+r^2 |V|)$ and an error complexity, in operator norm, of $(r |V|)^{1/2(2q+1)}\sigma_{r+1}$. In total, we obtain a decomposition procedure which is: exponentially exact with respect to $q$, at most quadratic in time with respect to $r$, at most linear in time with respect to graph parameters $|V|$ and $|E|$.

In the forward pass of the non-parametric aggregator, a graph with $d$-dimensional node features and $c$ classes will contribute a computational complexity of $\mathcal{O}(|V| c(d+r))$. Additionally, since the non-parametric aggregator is linear with respect to weights $W$ and parameter $\alpha$, gradient computation in the backward pass can re-use intermediaries found in the forward pass, potentially saving computation.

\section{Experiments}\label{sec:experi}

In an effort to show the power of feature aggregation for SSNC problems, our modeling effects will focus entirely on the aggregation matrix $\bm P$. No modifications are made to the original features $\bm X$ or the structure of the linear weight $\bm W$. As such, in our experiments we do not consider any model-specific augmentations such as dropout~\citep{Srivastava14},  batchnorm~\citep{Ioffe15}, or per-parameter optimizers (i.e. different learning rates for different layers). The design of $\bm P$ will have the following degrees of freedom:

\begin{itemize}
    \item \textbf{Matrix representation of network ($\bm M$):} We refer to any matrix $\bm M$ derived from algebraic manipulations of the adjacency of a network $\bm A$, to be a \emph{matrix representation} of the network. In particular we consider the following two representations:
    \begin{itemize}
        \item \emph{Adjacency}: This is simply an identity transformation on $\bm A$ with $\bm M = \bm A$.
        \item \emph{Laplacian}: This is $\bm M = \bm D - \bm A$ where 
        $\bm D$ is the row-sum degree matrix of $\bm A$. 
    \end{itemize}
    \item \textbf{Spectral truncation factor ($r$):} Given a truncation factor $r$, the spectral system $(\bm{U},\bm{\Lambda})$, resp. $(\bm{U},\bm{\Sigma}, \bm{V}^T)$, will be reduced to $(\bm{U}_{:r}, \bm{\Lambda}_{:r})$, resp. $(\bm{U}_{:r},\bm{\Sigma}_{:r},(\bm{V}_{:r})^T$), where the eigenvectors associated with the bottom $n-r$ eigenvalue magnitudes are dropped. In our experiments, spectral truncations  \underline{from 0 to 95\% in 5\% intervals} are considered.
    
    \item \textbf{Choice of kernel ($\mathcal{K}$):} In ordering our RKHS we select among the following kernels:
    \begin{itemize}
        \item \emph{Identity:} $K_{ij} = 1\{i=j\}$
        \item \emph{Linear (Outer product):} $\mathcal{K}(\sigma_i,\sigma_j) = \sigma_i\sigma_j$
        \item \emph{Compact Sobolev}: $\mathcal{K}(\sigma_i,\sigma_j) ={\rm min}(\sigma_i,\sigma_j)$
        \item \emph{Unbounded Sobolev:} $\mathcal{K}(\sigma_i,\sigma_j) = \exp(\gamma|\sigma_i-\sigma_j|)$
        \item \emph{Gaussian Radial Basis:} $\mathcal{K}(\sigma_i,\sigma_j) = \exp(\gamma|\sigma_i-\sigma_j|^2)$
    \end{itemize}
    Note, in the case of identity, the ``kernel" does not generate a continuous RKHS. For the last two kernels, the bandwidth parameter $\gamma\in\mathbb{R}_{+}$ can be determined on validation.
\end{itemize}

Note that, the choice of matrix representation $\bm M$ matters here insofar that it determines the ``modes" or partitions of the network with its left and right eigenvectors $(\bm U,\bm V)$. 

For our optimizer, we use the standard Adam optimizer~\citep{Kingma15} with weight decay. For simplicity, both parameter $\bm \alpha$ and weight matrix $\bm W$ share the same weight decay under Adam.

\subsection{SSNC Benchmarks}

Our methods are evaluated against common SSNC benchmarks. The Chameleon, Squirrel, and Actor benchmarks contain directed networks, while the other benchmarks contain undirected networks. More information on all benchmarks can be found in~\citet{Pei20}. All values are recorded using the \emph{balanced splits} defined in~\citet{Chien21}. Section~\ref{sec:eval:conv} provides a comprehensive analysis on the impact of splitting conventions. Although not covered in this paper, alternative benchmarks for simple spectral models can be found in Zhu and Koniusz~\cite{Zhu21}.

The following linear and kernel models are considered for evaluation: \textsc{Linear} ($\bm{XW}$), \textsc{Aggregated Linear} ($\bm{MXW}$), \textsc{Kernel} ($\bm{P}_{\mathcal{K}}\bm{XW}$), and \textsc{LR Kernel} ($\bm{P}_{\mathcal{K},r}\bm{XW}$). Model hyperparameters such as learning rate, weight decay, the specific aggregator $\bm P$ will be determined for each dataset using the mean accuracies of the validation splits. For completeness, we have also implemented a non-linear baseline which learns using only feature information $\bm X$. This model is a simple two-layer ReLU multi-layer perceptron \textsc{MLP2} ($\phi(\bm{XW^1}) \bm{W}^2$)
with hidden layer size determined on validation.

\begin{table}[t]
    \centering
    \resizebox{\textwidth}{!}{
    \begin{tabular}{l*{9}{c}}
        \toprule
        & {Cora} & {CiteSeer} & {PubMed} & {Chameleon} & {Squirrel} & {Actor} & {Cornell} & {Texas} & {Wisconsin} \\
        \midrule
        \textsc{MLP2} & \numwithuncertainty{77.8}{1.6} & \numwithuncertainty{77.2}{1.1} & \numwithuncertainty{88.2}{0.5} & \numwithuncertainty{48.5}{2.6} & \numwithuncertainty{34.8}{1.4} & \numwithuncertainty{40.3}{2.3} & \numwithuncertainty{86.1}{3.0} & \numwithuncertainty{91.7}{4.4} & \numwithuncertainty{95.0}{2.6} \\
        \textsc{Linear} & \numwithuncertainty{78.9}{2.0} & \numwithuncertainty{76.2}{1.2} & \numwithuncertainty{85.8}{0.4} & \numwithuncertainty{48.1}{3.2} & \numwithuncertainty{34.9}{1.4} & \numwithuncertainty{38.9}{1.2} & \numwithuncertainty{84.9}{5.6} & \numwithuncertainty{89.7}{3.8} & \numwithuncertainty{95.0}{3.8} \\
        \textsc{Agg. Linear} & \numwithuncertainty{84.0}{2.0} & \numwithuncertainty{73.9}{1.4} & \numwithuncertainty{82.6}{0.5} & \numwithuncertainty{79.0}{1.4} & \numwithuncertainty{78.0}{1.1} & \numwithuncertainty{32.4}{1.3} & \numwithuncertainty{67.8}{8.7} & \numwithuncertainty{86.8}{3.5} & \numwithuncertainty{83.8}{3.2} \\
        \textsc{Kernel} & \numwithuncertainty{88.6}{1.0} & \numwithuncertainty{81.1}{1.0} & \numwithuncertainty{89.4}{0.8} & \numwithuncertainty{78.7}{1.1} & \numwithuncertainty{76.0}{1.2} & \numwithuncertainty{32.2}{1.8} & \numwithuncertainty{83.3}{5.9} & \numwithuncertainty{88.2}{2.6} & \numwithuncertainty{92.1}{3.4} \\
        \textsc{LR Kernel} & {---} & {---} & {---} & \numwithuncertainty{79.4}{1.4} & \numwithuncertainty{76.8}{1.3} & \numwithuncertainty{32.3}{1.7} & {---} & {---} & {---} \\
        \midrule
        \textsc{GPRGNN}$^{*}$ & \numwithuncertainty{79.5}{0.4} & \numwithuncertainty{67.6}{0.4} & \numwithuncertainty{85.1}{0.1} & \numwithuncertainty{67.5}{0.4} & \numwithuncertainty{49.9}{0.5} & \numwithuncertainty{39.3}{0.3} & \numwithuncertainty{91.4}{0.7} & \numwithuncertainty{92.9}{0.6} & {NA} \\
        \textsc{SGC/ASGC}$^{*}$ & \numwithuncertainty{73.9}{2.5} & \numwithuncertainty{70.2}{1.0} & \numwithuncertainty{79.1}{1.0} & \numwithuncertainty{72.3}{0.9} & \numwithuncertainty{59.0}{1.0} & \numwithuncertainty{36.5}{0.8} &\numwithuncertainty{86.8}{3.6} &\numwithuncertainty{86.2}{3.1} & {NA} \\
        \textsc{JacobiConv}$^{*}$ & \numwithuncertainty{89.0}{0.5} & \numwithuncertainty{80.8}{0.8} & \numwithuncertainty{89.6}{0.4} & \numwithuncertainty{74.2}{1.0} & \numwithuncertainty{55.8}{0.6} & \numwithuncertainty{40.7}{1.0} & \numwithuncertainty{92.3}{2.8} & \numwithuncertainty{92.8}{2.0} & {NA} \\
        \textsc{ACMII-GCN} & \numwithuncertainty{89.0}{0.7} & \numwithuncertainty{81.8}{1.0} & \numwithuncertainty{90.7}{0.5} & \numwithuncertainty{68.4}{1.4} & \numwithuncertainty{54.5}{2.1} & \numwithuncertainty{41.8}{1.2} & \numwithuncertainty{95.9}{1.8} & \numwithuncertainty{95.1}{2.0} & \numwithuncertainty{96.6}{2.4} \\
        \bottomrule
    \end{tabular}
    }

    \medskip
    
    \caption{
     Performance: Mean test accuracy $\pm$ std. dev. over 10 data splits. Models include our own variations of ``Linear'' and ``Aggregated Linear'' GNNs, along with other state-of-the-art (SOTA) GNNs. Dashed entry in for \textsc{LR Kernel} signifies validated choice is the same as the full-rank \textsc{Kernel}. Performance is comparable between our simple GNNs and SOTA in some cases. Results for \textsc{GPRGNN}, \textsc{SGC/ASGC}. \textsc{JacobiConv} and \textsc{ACMII-GCN} are cited from~\cite{Chien21},~\cite{Chanpuriya22},~\cite{Wang22}, and~\cite{Luan22} respectively. Entries marked with `$*$' report 95\% confidence intervals.}
      \label{tab:balanced_results}
\end{table}

Our models and their results compared to other current SOTA methods can be found in Table~\ref{tab:balanced_results}. We note that, for almost all of the larger graph benchmarks, our models perform within uncertainty or better compared to SOTA. In particular for directed graphs like Chameleon and Squirrel, we see gains in accuracy as high as 5\% and 20\% over other SOTA methods.
A point of emphasis here is the relative simplicity of our models compared to the performance they attain. The absence of any post-model augmentations distinguishes our approach from the implementations of other competing SOTA spectral methods like \textsc{JacobiConv}~\citep{Wang22}.

A point of difficulty where the performance gap persists, is where 
the node response $\bm y$ is overwhelming described by its node information $\bm X$. Graphs with this property (Actor, Cornell, Texas, and Wisconsin) can be identified by the negative performance gap between \textsc{Linear} and \textsc{Aggregated Linear} as well as the SOTA-like performance of \textsc{MLP2}. Note that, even without using any graph information, \textsc{MLP2} is able to achieve SOTA within uncertainty on almost all of the $X$-dominated, network datasets. Furthermore, for the cases of Cornell, Texas, and Wisconsin, there is a possibility of running into sample size issues for graph based methods. With the exception of Actor, these datasets are only 100-200 nodes large (less than 1/10 the size of the other network benchmarks).

\begin{table}[t]
    \centering
    \renewcommand{\arraystretch}{1.3}
    {\small
    \begin{tabular}{lccccc}
        \toprule
        & max params.  & $n=300$ & $n=600$ & $n=1200$ & $n=1500$  \\ 
        \midrule
          \textsc{$X$-only oracle} & 0 & 
          $64.3 \pm 3.9$ & $66.2 \pm 2.9$ & $63.3 \pm 2.7$ & $64.6 \pm 1.4$\\
          \midrule 
          \textsc{Kernel} & 1512 & $75.0 \pm 3.5$ & $86.6 \pm 3.3$ & $94.5 \pm 1.2$ & $97.3 \pm 0.8$ \\
          \textsc{ACMII-GCN} & 102623 & $75.3 \pm 6.4$ & $89.5 \pm 3.1$ & $96.0 \pm 1.2$ & $97.7 \pm 0.8$ \\ 
        \bottomrule
    \end{tabular}
    }
    \medskip
    
    \caption{Simulation experiments on a three-class CSBM. Mean test accuracy and std. dev. of 10 runs are reported. \textsc{$X$-only oracle} is the accuracy associated with oracle classification on solely $X$. Maximum parameter counts for the two methods are also summarized. Relevant average degree $\Delta_n$ for the simulations are $\Delta_{300} =1.83$, $\Delta_{600} = 3.68$,  $\Delta_{1200} =7.58$, and $\Delta_{1500} = 9.44$.}
    \label{tab:csbm_results}
\end{table}

\subsection{CSBM Experiment}\label{sec:csbm}
To illustrate the effectiveness of nonparametric spectral learners, we performed an experiment on simulated CSBM data. We consider $C = 3$ classes and node features in $\reals^3$ ($\bm X \in \reals^{n \times 3}$) generated using \texttt{make\_blobs} function of \texttt{scikit-learn} package with cluster standard deviation of~10. This leads to a hard classification problem for an oracle that only knows $X$, with optimal Bayes accuracy of roughly $0.63$ (for large $n$). The SBM component has connection probabilities $B_{kk} = 0.015$ and  $B_{k\ell} = 0.02$ for $k \neq \ell$. We vary the number of nodes $n$ over $300, 600$, $1200$, and $1500$.

Table~\ref{tab:csbm_results} summarizes the results for our nonparametric learner (\textsc{Kernel}) and \textsc{ACMII-GCN} as a competing SOTA. Also shown is the average degree of the resulting networks. As $n$ increases the CSBM model becomes more informative, which is reflected in increased prediction accuracy. At the two ends of the SNR spectrum ($n=300$ and $n=1500$) the performance of the \textsc{Kernel} GNN and \textsc{ACMII-GCN} are very close, while there is a slight advantage for \textsc{ACMII-GCN} in the middle ($n=600$ and $n=1200$), though the two methods are still comparable due to the overlap of the wide uncertainty ranges.

What, however, is noteworthy is the significant effect of the spectral shaping in GNN performance: the \textsc{Kernel} GNN significantly improves the performance beyond the $X$-only oracle with very few parameters and at very low graph SNRs; for example, at $n=300$, where the parameter count is $312$ and the average degree is barely 2 (a very weak graph signal). The simplicity of the \textsc{Kernel} GNN allows us to exactly quantify the effect of nonparametric spectral learning since this is the only operation performed outside of applying the learned linear weights $\bm W$.

\subsection{Aggregation Ablation}

To understand the impact of the degrees of freedom defined for the aggregation matrix in section~\ref{sec:experi}, we conduct an ablation study on the three hyperparameters: matrix representation $\bm M$, truncation factor $r$, and the choice of kernel $\mathcal{K}$.

\paragraph{Matrix Representation ($\bm M$):} For this experiment we keep spectral truncation fixed at $0\%$ and choose the best kernel through validation splits. In other words, this experiment is conducted using the full-rank \textsc{Kernel} model with a best validated kernel fit to each dataset. In the experiment, we explore affects of fixing either $\bm M = \bm A$ or $\bm M = \bm D -\bm A$. 

\begin{figure}[b]
     \centering
     \includegraphics[width=0.775\linewidth]{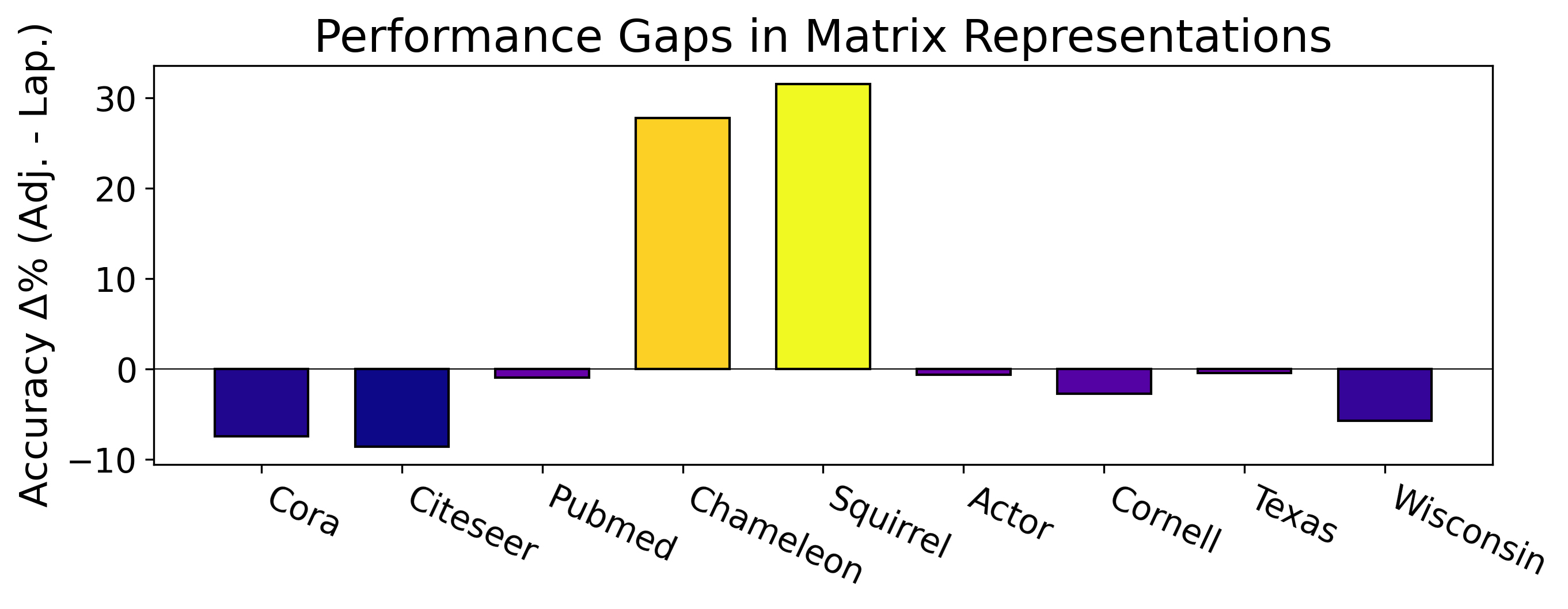}
     \caption{Accuracy comparison of the \textsc{Kernel} model for different graph representations $\bm A$ and $\bm D-\bm A$. Shown above is the signed accuracy difference between the adjacency and Laplacian representations. Best performing kernel was selected per dataset.} 
     \label{fig:graph_delt}
\end{figure}

Figure~\ref{fig:graph_delt} shows the accuracy change across datasets when using a Laplacian matrix representation $\bm D - \bm A$ rather than an adjacency matrix representation $\bm A$. 
As shown by the figure, directed graphs such as Chameleon and Squirrel show large benefits when using the adjacency matrix representation. Otherwise there seems to be a slight but persistent benefit in using the Laplacian representation for undirected datasets.

\begin{table}
    \centering
    \renewcommand{\arraystretch}{1.3}
    \resizebox{\columnwidth}{!}{
    \begin{tabular}{l|c|c|c|c|c|c|c|c|c}
        Kernel & Cora & CiteSeer & PubMed & Chameleon & Squirrel & Actor& Cornell & Texas & Wisconsin \\ \hline
        Identity &$78.8 \pm 2.7$ & $72.6 \pm 2.0$ & $81.6 \pm 0.9$ & $69.7 \pm 2.7$ & $44.9 \pm 2.9$ & $28.6 \pm 3.0$ & $60.4 \pm 8.1$ & $76.2 \pm 4.3$ & $71.6 \pm 5.7$\\
        Sob. Cmpct. & $75.1 \pm 1.9$ & $73.0 \pm 1.4$ & $88.5 \pm 0.4$ & $41.4 \pm 2.2$ & $33.2 \pm 1.1$ & $\bf{32.2 \pm 1.8}$ & $\bf{83.3 \pm 5.9}$ & $88.6 \pm 4.0$ & $\bf{92.1 \pm 3.4}$\\
        Linear & $81.1 \pm 2.0$ & $72.1 \pm 1.8$ & $82.3 \pm 1.0$ & $\bf{78.7 \pm 1.2}$ & $\bf{76.0 \pm 1.2}$ & $31.6 \pm 0.9$ & $66.5 \pm 6.1$ & $77.2 \pm 8.0$ & $81.3 \pm 4.8$\\
        Sob. Unbnd.& $88.8 \pm 0.8$ & $\bf{81.1 \pm 1.0}$ & $89.2 \pm 2.0$ & $54.5 \pm 6.4$ & $68.8 \pm 8.2$ & $30.7 \pm 1.0$ & $80.6 \pm 6.4$ & $\bf{88.2 \pm 2.6}$ & $90.4 \pm 5.6$\\
        Gauss. RBF& $\bf{88.6 \pm 1.0}$ & $80.3 \pm 1.9$ & $\bf{89.4 \pm 0.8}$ & $60.4 \pm 8.4$ & $71.3 \pm 4.4$ & $30.4 \pm 1.3$ & $79.4 \pm 5.3$ & $84.0 \pm 4.5$ & $85.8 \pm 4.7$ \\\hline
    \end{tabular}
    }
    
    \medskip
    
    \caption{Impact of the kernel choice on the  performance 
    of the full-rank \textsc{Kernel} model.
    Bold entries correspond to the model selected by validation. 
    }
    \label{tab:kern_results}
\end{table}

\textbf{Choice of Kernel ($\mathcal{K}$):} For this experiment we once again use the full-rank \textsc{Kernel} model. This time the matrix representation $\bm M$ is chosen through validation and the choice of kernel is varied across datasets. 
Table~\ref{tab:kern_results} shows performance results for the various choices of kernels. In Table~\ref{tab:kern_results}, we see a complicated dependence between kernel choice and the accuracy of node prediction. Although some results are within uncertainty, the dependence between kernel regularity and SSNC performance is not immediately clear. In the case of the Chameleon and Squirrel datasets, it is apparent that the wrong choice in kernel may lead to significant performance degradations (up to $\sim$30\%).

\begin{figure}[t]
     \centering
     \includegraphics[width=0.725\linewidth]{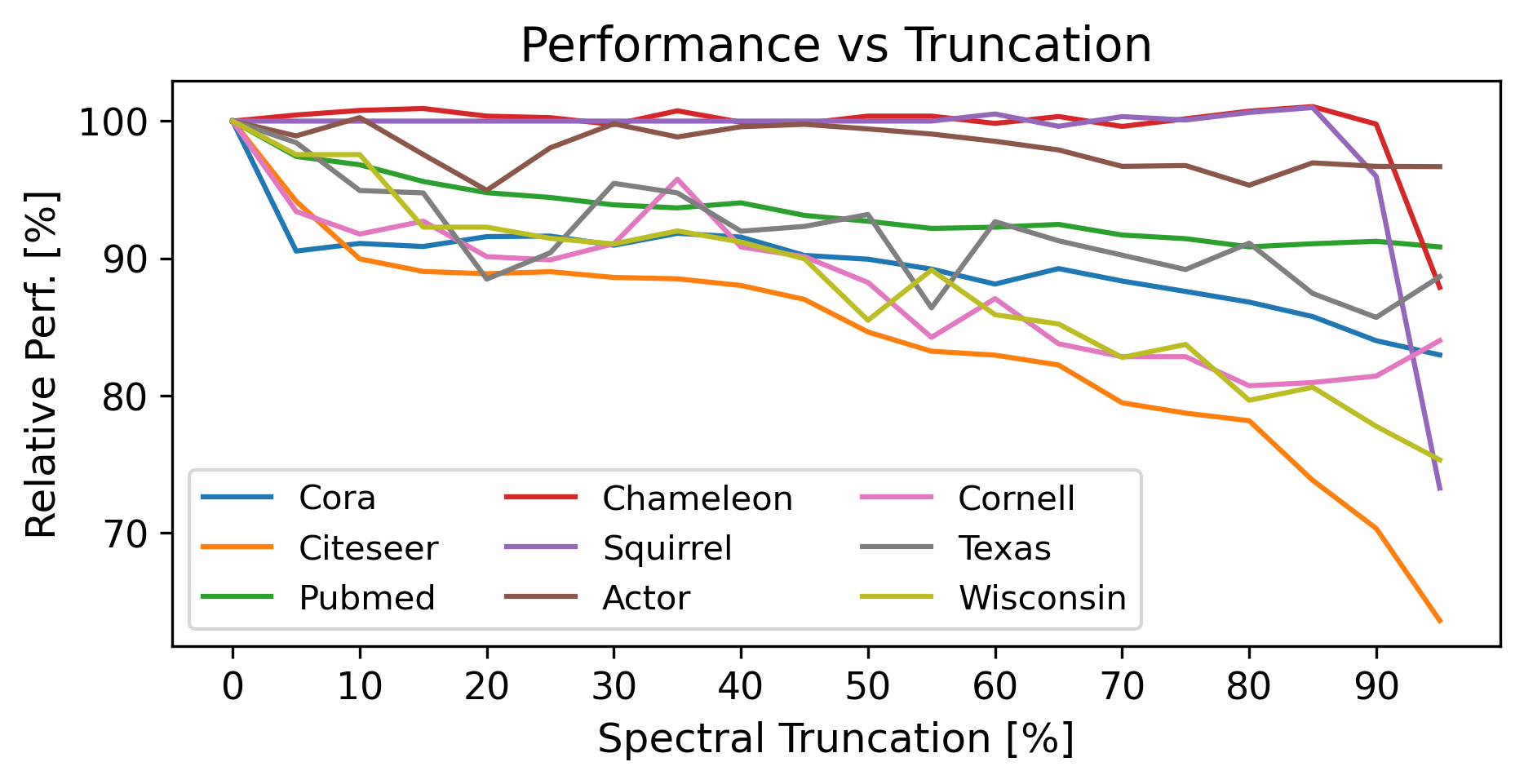}
     \caption{\textsc{LR Kernel} performance relative to the full-rank \textsc{Kernel} 
     for different 
     truncation factors $r$. Performance is seen to gradually decline on most datasets as the truncation factor $r$ decreases (that is truncation percentage increases). \textsc{LR Kernel} performance can also be seen to periodically increase above full-rank \textsc{Kernel} performance for the datasets Chameleon (red) and Squirrel (purple).} 
     \label{fig:truncations}
\end{figure}

\textbf{Spectral Truncation Factor ($r$):} For this experiment, both the matrix representation and the choice of kernel have been selected based on best validation with truncation factor $r$ fixed for the extent of each sub-experiment. Figure~\ref{fig:truncations} demonstrates the effect of truncation on performance and how it gradually degrades with the truncation percentage. The rate at which performance degrades seems dependent on the dataset, but most benchmarks retain $\sim$90\% performance even after a 40\% spectral truncation. In special cases like Squirrel and Chameleon, performance can be seen to increase at larger truncation values. 

\begin{figure}
     \centering
        \includegraphics[width=0.75\linewidth]{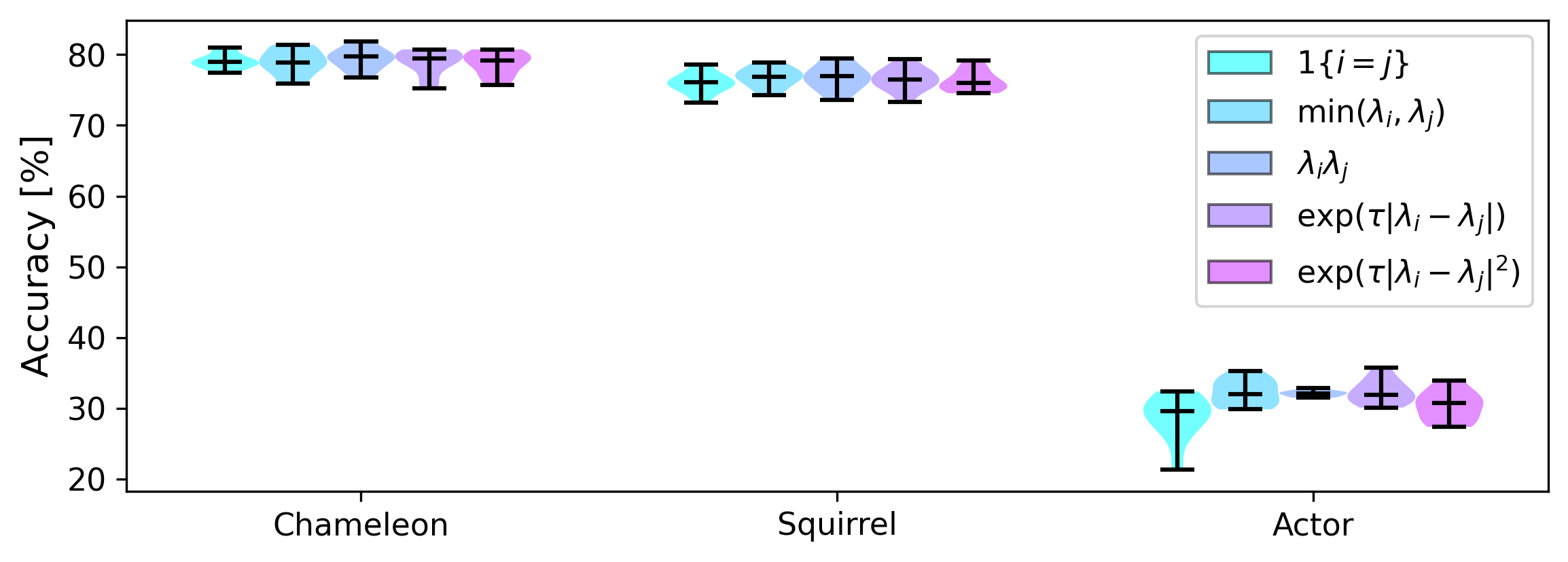}
         \caption{Performance homogenization
         achieved by \textsc{LR Kernel} model on 
         directed networks.}
     \label{fig:low_full_comp}
\end{figure}

\textbf{Alleviating Kernel Dependent Performance:} For this experiment, we explore the affects of kernel choice for the \textsc{LR Kernel} model. In particular, we focus on the performance impact of kernel choice for the directed dataset benchmarks. Rather than reporting mean accuracies, Figure~\ref{fig:low_full_comp} shows the full violin plot of test split performances for each kernel-dataset combination. We notice a \emph{homogenization} of results, where the choice of kernel is negligible to the overall SSNC performance. We stress however that this solution is partial, as the same order of homogenization is not observed for the other undirected datasets. Identifying relevant graph statistics which may describe this homogenization discrepancy is something which is left to future work.

\section{Changes in Evaluation Conventions}\label{sec:eval:conv}

The convention of using citation networks~\cite{Sen08} (Cora, Citeseer, Pubmed) in SSNC benchmarks was popularized by the graph embedding work of~\citet{Yang16}. \citet{Yang16} defined the ``sparse" train-test split of the citation datasets and their node masks were made publicly available. The sparse split fixed \emph{20 nodes per class for training} and \emph{1000 nodes total for testing}. These values were held constant across citation datasets, meaning larger networks likes Pubmed were left with a relatively low label rate of $\sim$5\%.

Quickly following was the semi-supervised work of~\citet{Kipf17} and~\citet{Velickovic18}. These follow-up papers defined a new ``public" split where \emph{500 previously unlabeled nodes} in the sparse split were now used for validation. In the respective code implementations of each paper, the additional labels were used for early stopping criteria and to determine the final model checkpoint.

\begin{figure}[b]
    \centering
    \includegraphics[width=0.9\linewidth]{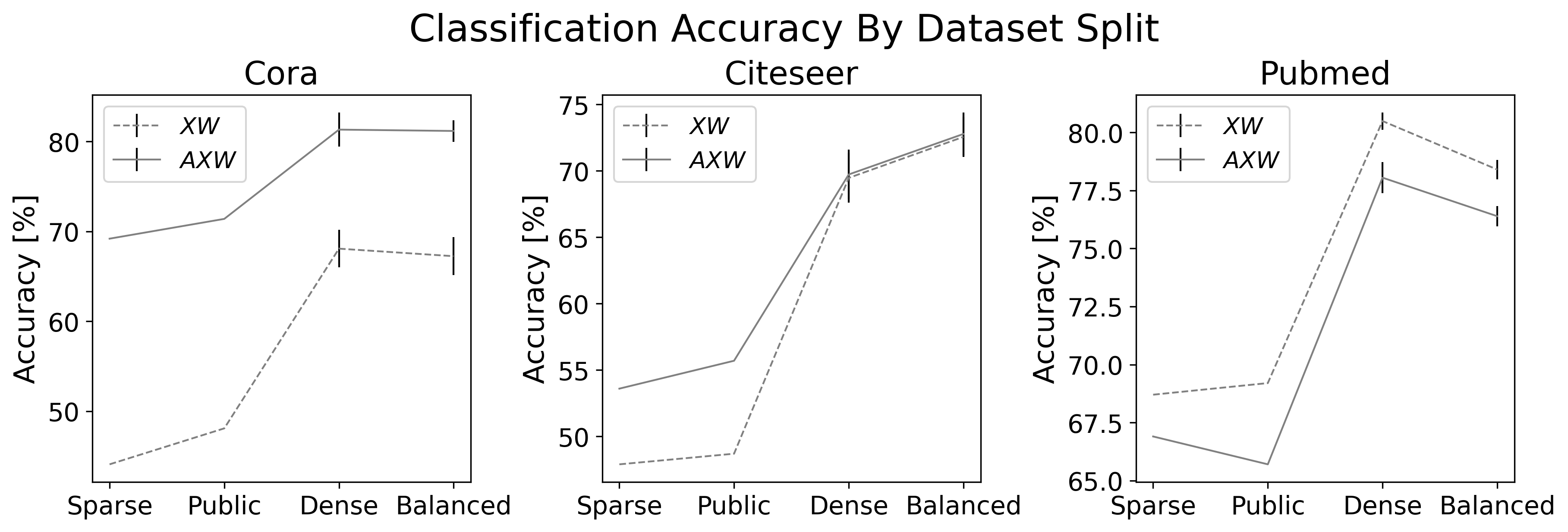}
    \caption{Accuracy results and uncertainties on the citation datasets using different splits with linear models $\bm{XW}$ and $\bm{AXW}$. ``Public" refers to the split introduced by~\citet{Kipf17}. Both ``Sparse" and ``Public" are single splits, so 
    one cannot 
    associate uncertainty to them.}
    \label{fig:citation}
\end{figure}

Introduced later was the ``dense'' split by~\citet{Pei20}, where train, validation, and test were now fractions of the whole graph, set to \emph{60\%-20\%-20\% respectively}. This paper also popularized two new benchmark datasets, the WebKB dataset~\cite{Craven98} (Cornell, Texas, Wisconsin) and the Wikipedia animal page-page networks~\cite{Rozemberczki21} (Chameleon, Squirrel). 

Most recently a ``balanced" split was proposed by~\citet{Chien21}. This is a class-balanced split where, \emph{for each class in a network}, a 60\%-20\%-20\% mask is made with then each class mask being collected into a final, aggregate train-validation-test split. Both the balanced split and the datasets tested in Section~\ref{sec:experi} are commonplace benchmarking practices for current SSNC papers~\citep{Luan22,Wang22}.

\begin{figure}[t]
    \centering
    \includegraphics[width=0.75\linewidth, height=4.5cm]{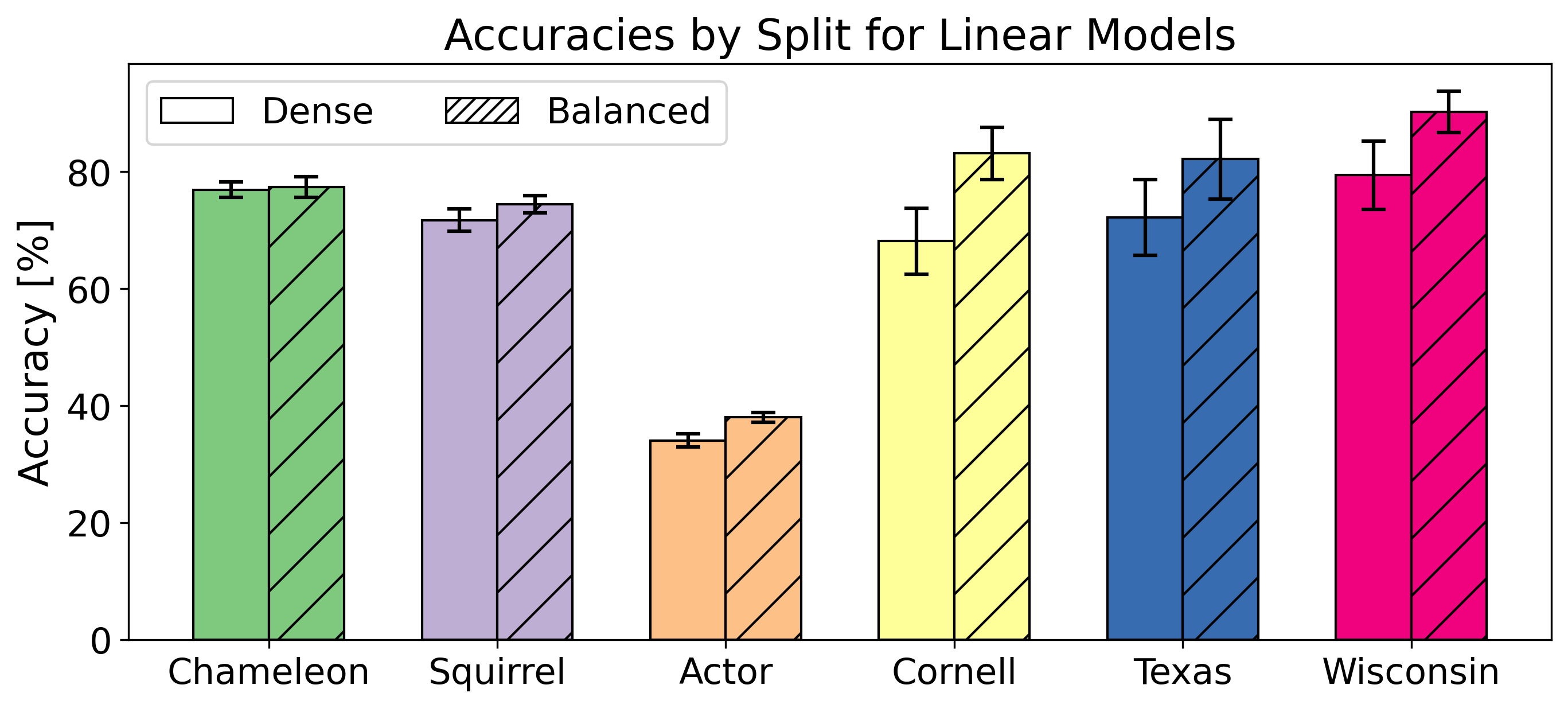}
    \caption{Accuracy results on datasets introduced by~\citet{Pei20}. ``Dense" refers to the original split while ``Balanced" refers to the split introduced by~\citet{Chien21}. Test results and uncertainties are evaluated using models $\bm{XW}$ and $\bm{AXW}$. Results shown are for method with best validation.}
    \label{fig:linear_best}
\end{figure}

\subsection{Comparing Split Performances}

Provided in Figures~\ref{fig:citation}-\ref{fig:linear_best} are visualizations on the impacts of different evaluation techniques on simple linear models ($\bm{XW}$ and $\bm{AXW}$). 
To keep things comparable to the sparse split, where no validation set exists, both the learning rate 
($10^{-3}$) and the weight decay (0.0) were set to be fixed for the Adam optimizer. 
Despite this lack of tuning, the best of these models, per dataset, achieve roughly $>$85\% relative performance when compared to SOTA SSNC methods. The high-end of this performance can be seen in the Squirrel column of Figure~\ref{fig:linear_best}, where mean accuracy of the best linear model is 77.3\%.

New GNN architectures which make use of the recent, balanced split may also experience an analogous performance bump relative to any older models tested before the split was introduced. In the worst case, this may lead to an overstatement in new modeling contributions and has the potential downside of muddying the signal of what makes for a successful and efficient GNN architecture in SSNC experiments. For this reason, we believe it is important to be clear on the impact of splitting conventions and how they contribute to recent performance upticks in SSNC benchmarking.

\section{Conclusions}

We have shown how classically-inspired, non-parametric techniques can be used to match, and sometimes exceed, previous spectral and non-linear GNN approaches. Our methods make no use of post-model augmentations, such as dropout~\citep{Srivastava14} or batchnorm~\citep{Ioffe15}, and allow for a clean theoretical analysis in future work. 

Empirically, we explored and ablated pertinent hyperparameters to the spectral kernel model and have shown the various dependences between parameters across different datasets.
On the aspect of low-rank kernel models, we have shown how spectral truncation can homogenize response outcomes for different kernel choices. Additionally for low-rank models, we have shown how performance decline is gradual with increases in spectral truncation, pointing to practical speed-ups for non-parametric kernel implementations.

On the aspect of testing conventions, we looked at how evaluation has changed for SSNC tasks since the first introduction of popular citation datasets~\citep{Sen08}. We have shown how the class-balanced split can produce improvements in performance outside of what is expected by uncertainty.

In summary, non-parametric kernel aggregators provide a simple yet effective means of recovering unobserved labels in SSNC tasks. As our implementations are free from post-model augmentations, we expect future theoretical insights obtained for low rank kernel aggregators to be closely reproduced in experimental settings such as those seen in Section~\ref{sec:experi}. Future work may further develop these insights, adding to the list of favorable properties for non-parametric kernel aggregators.

\bibliography{iclr2024_conference}

\begin{thebibliography}{24}
\providecommand{\natexlab}[1]{#1}
\providecommand{\url}[1]{\texttt{#1}}
\expandafter\ifx\csname urlstyle\endcsname\relax
  \providecommand{\doi}[1]{doi: #1}\else
  \providecommand{\doi}{doi: \begingroup \urlstyle{rm}\Url}\fi

\bibitem[Seeger(2002)]{Seeger02}
Matthias Seeger.
\newblock Learning with labeled and unlabeled data.
\newblock Technical report, Institute for Adaptive and Neural Computation, University of Edinburgh, 2002.

\bibitem[Belkin et~al.(2006)Belkin, Niyogi, and Sindhwani]{Belkin06}
Mikhail Belkin, Partha Niyogi, and Vikas Sindhwani.
\newblock Manifold regularization: A geometric framework for learning from labeled and unlabeled examples.
\newblock \emph{J. Mach. Learn. Res.}, 7:\penalty0 2399–2434, Dec. 2006.

\bibitem[Zhu et~al.(2003)Zhu, Ghahramani, and Lafferty]{Zhu03}
Xiaojin Zhu, Zoubin Ghahramani, and John~D. Lafferty.
\newblock Semi-supervised learning using gaussian fields and harmonic functions.
\newblock In \emph{International Conference on Machine Learning}, 2003.

\bibitem[Scarselli et~al.(2009)Scarselli, Gori, Tsoi, Hagenbuchner, and Monfardini]{Scarselli09}
Franco Scarselli, Marco Gori, Ah~Chung Tsoi, Markus Hagenbuchner, and Gabriele Monfardini.
\newblock The graph neural network model.
\newblock \emph{IEEE Transactions on Neural Networks}, 20\penalty0 (1):\penalty0 61--80, 2009.

\bibitem[Veličković et~al.(2018)Veličković, Cucurull, Casanova, Romero, Liò, and Bengio]{Velickovic18}
Petar Veličković, Guillem Cucurull, Arantxa Casanova, Adriana Romero, Pietro Liò, and Yoshua Bengio.
\newblock Graph attention networks.
\newblock In \emph{International Conference on Learning Representations}, 2018.

\bibitem[Chien et~al.(2021)Chien, Peng, Li, and Milenkovic]{Chien21}
Eli Chien, Jianhao Peng, Pan Li, and Olgica Milenkovic.
\newblock Adaptive universal generalized pagerank graph neural network.
\newblock In \emph{International Conference on Learning Representations}, 2021.

\bibitem[Luan et~al.(2022)Luan, Hua, Lu, Zhu, Zhao, Zhang, Chang, and Precup]{Luan22}
Sitao Luan, Chenqing Hua, Qincheng Lu, Jiaqi Zhu, Mingde Zhao, Shuyuan Zhang, Xiao-Wen Chang, and optdoina Precup.
\newblock Revisiting heterophily for graph neural networks.
\newblock In \emph{Advances in Neural Information Processing Systems}, volume~35, pages 1362--1375, 2022.

\bibitem[Defferrard et~al.(2016)Defferrard, Bresson, and Vandergheynst]{Defferrard16}
Micha\"{e}l Defferrard, Xavier Bresson, and Pierre Vandergheynst.
\newblock Convolutional neural networks on graphs with fast localized spectral filtering.
\newblock In \emph{Proceedings of the 30th International Conference on Neural Information Processing Systems}, NIPS'16, page 3844–3852, Red Hook, NY, USA, 2016. Curran Associates Inc.

\bibitem[Wang and Zhang(2022)]{Wang22}
Xiyuan Wang and Muhan Zhang.
\newblock How powerful are spectral graph neural networks.
\newblock In \emph{Proceedings of the 39th International Conference on Machine Learning}, volume 162 of \emph{Proceedings of Machine Learning Research}, pages 23341--23362, 17--23 Jul 2022.

\bibitem[Rozemberczki et~al.(2021)Rozemberczki, Allen, and Sarkar]{Rozemberczki21}
Benedek Rozemberczki, Carl Allen, and Rik Sarkar.
\newblock {Multi-Scale attributed node embedding}.
\newblock \emph{Journal of Complex Networks}, 9\penalty0 (2):\penalty0 cnab014, 05 2021.

\bibitem[Sch{\"o}lkopf et~al.(2001)Sch{\"o}lkopf, Herbrich, and Smola]{Scholkopf01}
Bernhard Sch{\"o}lkopf, Ralf Herbrich, and Alex~J. Smola.
\newblock A generalized representer theorem.
\newblock In \emph{Computational Learning Theory}, pages 416--426, Berlin, Heidelberg, 2001. Springer Berlin Heidelberg.

\bibitem[Deshpande et~al.(2018)Deshpande, Sen, Montanari, and Mossel]{Deshpande18}
Yash Deshpande, Subhabrata Sen, Andrea Montanari, and Elchanan Mossel.
\newblock Contextual stochastic block models.
\newblock In S.~Bengio, H.~Wallach, H.~Larochelle, K.~Grauman, N.~Cesa-Bianchi, and R.~Garnett, editors, \emph{Advances in Neural Information Processing Systems}, volume~31. Curran Associates, Inc., 2018.

\bibitem[Young and Scheinerman(2007)]{Young07}
Stephen~J. Young and Edward~R. Scheinerman.
\newblock Random dot product graph models for social networks.
\newblock In \emph{Algorithms and Models for the Web-Graph}, pages 138--149, Berlin, Heidelberg, 2007. Springer Berlin Heidelberg.

\bibitem[Halko et~al.(2011)Halko, Martinsson, and Tropp]{Halko11}
N.~Halko, P.~G. Martinsson, and J.~A. Tropp.
\newblock Finding structure with randomness: Probabilistic algorithms for constructing approximate matrix decompositions.
\newblock \emph{SIAM Review}, 53\penalty0 (2):\penalty0 217--288, 2011.

\bibitem[Srivastava et~al.(2014)Srivastava, Hinton, Krizhevsky, Sutskever, and Salakhutdinov]{Srivastava14}
Nitish Srivastava, Geoffrey Hinton, Alex Krizhevsky, Ilya Sutskever, and Ruslan Salakhutdinov.
\newblock Dropout: A simple way to prevent neural networks from overfitting.
\newblock \emph{Journal of Machine Learning Research}, 15\penalty0 (56):\penalty0 1929--1958, 2014.

\bibitem[Ioffe and Szegedy(2015)]{Ioffe15}
Sergey Ioffe and Christian Szegedy.
\newblock Batch normalization: Accelerating deep network training by reducing internal covariate shift.
\newblock In \emph{Proceedings of the 32nd International Conference on Machine Learning}, volume~37 of \emph{Proceedings of Machine Learning Research}, pages 448--456, Lille, France, 07--09 Jul 2015.

\bibitem[Kingma and Ba(2015)]{Kingma15}
Diederik~P. Kingma and Jimmy Ba.
\newblock Adam: {A} method for stochastic optimization.
\newblock In \emph{3rd International Conference on Learning Representations, {ICLR} 2015, San Diego, CA, USA, May 7-9, 2015, Conference Track Proceedings}, 2015.

\bibitem[Pei et~al.(2020)Pei, Wei, Chang, Lei, and Yang]{Pei20}
Hongbin Pei, Bingzhe Wei, Kevin Chen-Chuan Chang, Yu~Lei, and Bo~Yang.
\newblock Geom-gcn: Geometric graph convolutional networks.
\newblock In \emph{International Conference on Learning Representations}, 2020.

\bibitem[Zhu and Koniusz(2021)]{Zhu21}
Hao Zhu and Piotr Koniusz.
\newblock Simple spectral graph convolution.
\newblock In \emph{International Conference on Learning Representations}, 2021.

\bibitem[Chanpuriya and Musco(2022)]{Chanpuriya22}
Sudhanshu Chanpuriya and Cameron~N Musco.
\newblock Simplified graph convolution with heterophily.
\newblock In Alice~H. Oh, Alekh Agarwal, Danielle Belgrave, and Kyunghyun Cho, editors, \emph{Advances in Neural Information Processing Systems}, 2022.

\bibitem[Sen et~al.(2008)Sen, Namata, Bilgic, Getoor, Galligher, and Eliassi-Rad]{Sen08}
Prithviraj Sen, Galileo Namata, Mustafa Bilgic, Lise Getoor, Brian Galligher, and Tina Eliassi-Rad.
\newblock Collective classification in network data.
\newblock \emph{AI Magazine}, 29\penalty0 (3):\penalty0 93, Sep. 2008.

\bibitem[Yang et~al.(2016)Yang, Cohen, and Salakhutdinov]{Yang16}
Zhilin Yang, William~W. Cohen, and Ruslan Salakhutdinov.
\newblock Revisiting semi-supervised learning with graph embeddings.
\newblock In \emph{Proceedings of the 33rd International Conference on International Conference on Machine Learning - Volume 48}, ICML'16, page 40–48, 2016.

\bibitem[Kipf and Welling(2017)]{Kipf17}
Thomas~N. Kipf and Max Welling.
\newblock Semi-supervised classification with graph convolutional networks.
\newblock In \emph{International Conference on Learning Representations}, 2017.

\bibitem[Craven et~al.(1998)Craven, DiPasquo, Freitag, McCallum, Mitchell, Nigam, and Slattery]{Craven98}
Mark Craven, Dan DiPasquo, Dayne Freitag, Andrew McCallum, Tom Mitchell, Kamal Nigam, and Se\'{a}n Slattery.
\newblock Learning to extract symbolic knowledge from the world wide web.
\newblock AAAI '98/IAAI '98, page 509–516, USA, 1998.

\end{thebibliography}
\bibliographystyle{unsrtnat}

\end{document}